\documentclass[sigconf]{acmart}
\newif\ifnotnewacm
\notnewacmfalse

\newif\ifheadnice
\headnicefalse

\begin{CJK}{UTF8}{gbsn}

\newtheorem{algorithm}{算法}
\newtheorem{theorem}{定理}[section]  
\newtheorem{definition}{定义}

\newtheorem{lemma}{引理}

\newtheorem{conclusion}{结论}

\renewcommand{\abstractname}{摘要}  
\renewcommand{\refname}{参考文献}   

\title{***}
\author{***footnote{电子邮件:**}\\[2ex]
*** \\[2ex]
}
\date{20XX年X月}

\maketitle

\tableofcontents
\newpage
在此输入正文，中英文均可。

\songti 
\fangsong 
\biaosong
\heiti
\kaishu

\end{CJK}

\newif\ifeg
\egfalse

\def\pku{\superscript{$\mathbb{P}$}}

\def\hcs{\superscript{$\mathbb{K}$}}
\def\pitt{\superscript{$\mathbb{T}$}}

\def\sys{\textsc{SynCheck}\xspace}

\definecolor{pkured}{HTML}{BD362F}
\newcommand{\bo}[1]{{\color{pkured}[BO: #1]}}
\newcommand{\chen}[1]{{\color{teal}[CHEN: #1]}}

\setlist[itemize]{leftmargin=10pt,itemsep=0pt,parsep=0pt}
\setlist[enumerate]{leftmargin=0pt}
\usepackage{setspace}
\usepackage{adjustbox}
\usepackage{tcolorbox}
\usepackage{makecell}
\theoremstyle{definition}
\newtheorem{definition}{Definition}
\newtheorem{theorem}{Theorem}
\newtheorem{lemma}{Lemma}
\newtheorem{conclusion}{Conclusion}
\usepackage{pifont}
\usepackage{multirow}
\DeclareMathOperator{\TV}{TV} 
\usepackage{amsmath}
\usepackage{subcaption}
\newcommand{\tightsubfigures}{\captionsetup[subfigure]{skip=-3pt}}

\begin{document}
\fancyhead{} 

\title{Data Can Speak for Itself: Quality-guided Utilization \\ of Wireless Synthetic Data}

\author{Chen Gong\pku, Bo Liang\pku, Wei Gao\pitt, Chenren Xu\pku\hcs\superscript{\Letter}}

\authornote{
    \Letter: chenren@pku.edu.cn
}

\affiliation{
    \begin{tabular}{c}
        {\pku}School of Computer Science, Peking University; {\pitt}University of Pittsburgh \\
        {\hcs}Key Laboratory of High Confidence Software Technologies, Ministry of Education (PKU)
    \end{tabular}
    \vspace{1mm}
}

\def\authors{Chen Gong, Bo Liang, Wei Gao, Chenren Xu}

\begin{abstract}
Generative models have gained significant attention for their ability to produce realistic synthetic data that supplements the \textit{quantity} of real-world datasets. While recent studies show performance improvements in wireless sensing tasks by incorporating \textit{all} synthetic data into training sets, the \textit{quality} of synthetic data remains unpredictable and the resulting performance gains are not guaranteed. To address this gap, we propose tractable and generalizable metrics to quantify quality attributes of synthetic data—affinity and diversity. Our assessment reveals prevalent affinity limitation in current wireless synthetic data, leading to mislabeled data and degraded task performance. We attribute the quality limitation to generative models' lack of awareness of untrained conditions and domain-specific processing. To mitigate these issues, we introduce \sys, a quality-guided synthetic data utilization scheme that refines synthetic data quality during task model training.
Our evaluation demonstrates that \sys consistently outperforms quality-oblivious utilization of synthetic data, and achieves 4.3\% performance improvement even when the previous utilization degrades performance by 13.4\%.
\end{abstract}



\acmYear{2025}\copyrightyear{2025}
\setcopyright{acmlicensed}
\acmConference[MobiSys '25]{The 23rd Annual International Conference on Mobile Systems, Applications and Services}{June 23--27, 2025}{Anaheim, CA, USA}
\acmBooktitle{The 23rd Annual International Conference on Mobile Systems, Applications and Services (MobiSys '25), June 23--27, 2025, Anaheim, CA, USA}
\acmDOI{10.1145/3711875.3729123}
\acmISBN{979-8-4007-1453-5/25/06}

\maketitle

\section{Introduction}
\label{sec:intro}

\begin{figure*}[t]
    \centering
    \includegraphics[width=\linewidth]{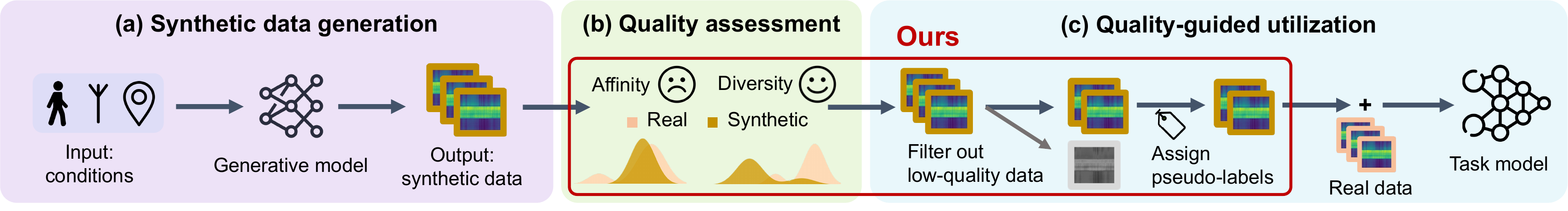}
    \vspace{-4mm}
    \caption{
    Our quality assessment and quality-guided utilization of wireless synthetic data.
    \textmd{Generative models produce synthetic data from conditions to supplement data \textit{quantity} for wireless applications. Compared to previous quality-oblivious utilization using \textit{all} synthetic data with \textit{conditions as labels}, we assess synthetic data quality, reveal its affinity limitation, and propose a quality-guided utilization scheme incorporating \textit{filtered} synthetic samples with \textit{assigned pseudo-labels} for better data quality and task performance.}}
    \vspace{-2mm}
    \label{fig:intro_fig1}
\end{figure*}

Recent technical advances in generative AI models have shed light on the possibility to generate realistic synthetic data~\cite{ddpm,gan_intro,vqvae}. One important variant is the conditional generative models that can guide the generative process with input class labels and produce output from different specified types~\cite{Mirza2014ConditionalGA}, such as the generation of handwritten words based on text condition to facilitate tasks like word recognition~\cite{Kang2020GANwritingCG}.
Following this trend, the wireless research community, long facing data scarcity, has adopted various conditional generative models~\cite{csigan,crossfreq_gan,fido,rf-diffusion,rf-genesis} to augment the availability of training data with synthetic samples produced with task-specific conditions, such as human action and antenna location in part (a) of \figref{fig:intro_fig1}. The \textit{labeled synthetic data} has proved effective to improve the performance of data-hungry wireless applications such as action recognition~\cite{crossgr} and indoor localization~\cite{fidora}.

While most work in data synthesis for data scarcity alleviation approves its success in supplementing \textit{quantity}~\cite{singh2021medical,chen2023cross,LearnSense}, prior studies have highlighted a remaining deficiency of synthetic data \textit{quality} in capturing domain-specific knowledge, such as that required in medical applications~\cite{adams2023does,chambon2022adapting}. To bridge this gap, human expert feedback is frequently employed for calibration~\cite{sun2024aligning}.   However, wireless signals are not understandable by humans, making human-driven quality evaluation impractical. 
Given that the nonselective utilization of all synthetic data has been shown to yield suboptimal task performance~\cite{faithful}, we ask a question in this paper: \textit{How to evaluate the quality of wireless synthetic data and how can the quality assessment guide the utilization of synthetic data?}


Answering this question is not easy, as there are several technique challenges of conducting and leveraging the synthetic data quality assessment in the wireless domain. The first challenge is to \textbf{define tractable and generalizable quality quantification metrics (C1)}. In established synthetic data quality assessment~\cite{chen2021synthetic,faithful}, affinity and diversity are recognized key attributes: \textit{Affinity captures the realism of synthetic data in its similarity to real data, while diversity emphasizes the data variability to extend the original distribution of real data}. Current quantification metrics for the two attributes often rely on empirical comparisons of synthetic data generated from the same training set~\cite{lucic2018gans,stein2024exposing}, which is problematic for wireless generative models tailored to specific hardware and signal properties~\cite{uwb_fi,rf-genesis} with customized datasets, making it difficult to perform unified comparisons on common datasets. Therefore, wireless synthetic data quality assessment requires generalizable metrics, with tractable theoretical support, to enable fair cross-dataset comparisons.

The second challenge is to evaluate \textbf{heterogeneous data generation techniques (C2)}. Wireless researchers have developed advanced generation techniques across diverse applications~\cite{ouyang2024mmbind,ouyang2024llmsense,dai2024advancing}. Given the breadth of research, systematic evaluation within a reasonable computational budget is essential to draw reliable and universal conclusions about the quality of wireless synthetic data.  

To address the challenges of tractable and generalizable quality assessment (\textbf{C1}), we propose quality quantification metrics with both \textit{theory justification and cross-dataset comparability}.  Our approach incorporates Bayesian analysis and performance indicators that can be consistently referenced across datasets, effectively overcoming the limitations of previous task-oriented performance metrics~\cite{good_gan} in theoretical support and generalization. 
Specifically, we interpret affinity and diversity as conditional probability distribution between samples and labels, and establish the connection between these attributes and task performance across different training sets. Additionally, we propose to employ the overall classification confidence distribution as performance indicator~\cite{bartlett2017spectrally,elsayed2018large}, where training set serves as reference for each dataset. This approach enables tractable and generalizable quality assessment for each wireless generative model. 

To efficiently account for the diverse generation techniques of wireless synthetic data (\textbf{C2}), we categorize the synthetic data according to its source and target domains of generation, and conduct a \textit{rigorous and systematic quality quantification of synthetic data} from representative generative models with our metrics. Our quality assessment reveals that current wireless synthetic data suffer from prevalent \textit{limited affinity}, yet possess \textit{good diversity} for classes in the training set, as shown by the part (b) of \figref{fig:intro_fig1}. Consequently, prior quality-oblivious utilization of synthetic data with limited affinity results in training data with incorrect labels. The mislabeled data can distort training process and degrade performance~\cite{aum,zhang2021understanding}, and we observe performance degradation when trained with synthetic data deficient in both affinity and diversity in \secref{sec:amount_impact}. 

Having identified the quality deficiencies, we propose a \textbf{quality-guided utilization scheme} designed to \textit{enhance synthetic data affinity while preserving its diversity}. This solution addresses a critical gap in existing work within the wireless domain, which has focused on advancing data generation techniques but overlooks effective utilization of synthetic data for downstream tasks~\cite{rf-diffusion, mmgpe}. Our key rationale is to treat synthetic data as unlabeled and real data as labeled, combining both data sources with a \textit{semi-supervised learning} framework. Specifically, our approach, \sys, filters out synthetic samples with low affinity and assigns pseudo-labels to remaining samples during iterative model training, as illustrated in part (c) of \figref{fig:intro_fig1}. This utilization scheme is general post-processing step without modifying the training or inference of generative models, making it adaptable to various generation processes.



We implement \sys and conduct extensive experiments across various categories of wireless synthetic data. Our results show that \sys improves task performance by 12.9\% to 4.3\%, while performance gain with quality-oblivious utilization of synthetic data is 4.1\%. Even in worst-case scenarios, \sys yields a 4.3\% improvement, while previous utilization degrades performance by 13.4\%.
Furthermore, synthetic data filtered by \sys exhibits enhanced affinity and comparable diversity to unfiltered raw data. The alignment between task performance and our quality metrics validates the effectiveness of our quality metrics and the necessity of quality-guided utilization of synthetic data\footnote{Codes are available at \url{https://github.com/MobiSys25AE/SynCheck}.}. 


Our contributions in this paper are summarized as follows:
\begin{itemize}
    \item We introduce the first tractable and generalizable metrics for quantifying affinity and diversity in wireless synthetic data. Our data quality evaluation of representative generative models reveals limited affinity but good diversity.
    \item We introduce \sys, to the best of our knowledge,the first quality-guided scheme for wireless synthetic data utilization, improving task performance with better data affinity. 
    \item We conduct extensive experiments and show that \sys boosts task performance compared to  nonselective utilization of real and synthetic data.
\end{itemize}

\section{Background}

\begin{table*}[t]
    \centering
    \caption{Comparison of existing quantification metrics on synthetic data quality.}
    \vspace{-5pt}
    \centering
    \label{tab:quality}
    \resizebox{0.9\linewidth}{!}{
    \begin{tabular}{ccccccc}
        \toprule
        \multirow{2}{*}{Literature} & \multicolumn{2}{c}{Quality Metrics} & \multirow{2}{*}{Quantification Rationale} & \multicolumn{2}{c}{Properties}\\ 
        & Affinity & Diversity &  & Tractable & Generalizable \\ 
        \midrule
        QAGAN~\cite{QAGAN} & SSIM & FID & \makecell{Visual similarity} & \ding{55} & \ding{51}\\ 
        
        GoodGAN~\cite{good_gan} & \makecell{TRTS accuracy} & \makecell{TSTR accuracy} & Task accuracy & \ding{55} & \ding{55}\\
        
        CAS~\cite{cas} & (\textit{missing}) &  \makecell{TSTR w. $p_{\theta}(y|x)$} & \makecell{Bayesian analysis} & \ding{51} & \ding{55}\\
        
        Faithful~\cite{faithful} & $\alpha$-precision &  $\beta$-recall & \makecell{Hypersphere distribution} & \ding{55} & \ding{51}\\
        \midrule
        Ours & \makecell{TR margin w. $p_{\theta}(x|y)$} &  \makecell{TS margin w. $p_{\theta}(y|x)$} & \makecell{Task confidence w. Bayesian analysis} & \ding{51} & \ding{51}\\
        \bottomrule
    \end{tabular}
    }
\end{table*}

\subsection{Wireless Data Generation}\label{ssec: wireless data generation} 
The scarcity of high-quality large-scale training datasets has long been a challenge in wireless community \cite{cai2020teaching,sensing_survey}. 
To address this limitation and alleviate the burden of data collection, data generation techniques have been developed to expand quantity through augmentation or synthesis~\cite{generation_survey1}.


\nosection{Data Augmentation} Augmentation techniques extend existing data by modifying the outer representations while maintaining the inner semantics, informed by domain knowledge.
It is widely used across various domains and proves effective in boosting performance.
In the wireless domain, data augmentation models the variation of channels and transceivers. 
Examples include multi-scale frequency resolution and motion-aware random erasing and shifting~\cite{rfboost}, random phase shifts and amplitude fluctuations~\cite{aug2}, and random circular shifts and re-scaling~\cite{aug3}. 


\nosection{Data Synthesis} Synthesis techniques generate new data either by adhering to physical laws~\cite{ray_tracing,nerf2,rf-genesis} or through deep learning models~\cite{rf-eats}. The former approach typically involves environment modeling and simulation based on physical rules to replicate real data, while the latter learns the data distribution of the training set and synthesizes new data that follows this underlying distribution. Given the complexity and over-idealization of accurate physical modeling, which is often limited to a single scenario, this study focuses on \textit{data-driven generative models} and \textit{aims to advance our understanding of the quality of wireless synthetic data and enhance its utilization to improve task performance}.

\subsection{Synthetic Data Quality Assessment}

\label{ssec:generative_eval} 

The quality of synthetic data has been extensively studied in generative model research~\cite{survey_eval1,survey_eval2}. However, existing metrics fail to simultaneously achieve tractability for theoretical grounding and generalizability for task- and dataset-agnostic evaluation, as they either rely on visual heuristics or are designed for controlled tests on the same dataset. To better understand these limitations, we summarize relevant work on affinity and diversity metrics in \tabref{tab:quality}.

\nosection{Deficiencies of Existing Metric Designs} \textit{Visual similarity} metrics are widely adopted for synthetic image evaluation, including per-sample structural similarity (SSIM)~\cite{ssim_gan} and overall distribution diversity (FID)~\cite{fid}. However, these similarity-based metrics rely on heuristic assumptions and fail to align with human perception~\cite{lucic2018gans, stein2024exposing}, limiting their theoretical tractability. 
\textit{Task-oriented} metrics evaluate real-world task performance in \textit{train-real-test-synthetic} (TRTS) and \textit{train-synthetic-test-real} (TSTR) setups~\cite{good_gan}. The former trains a model on real data and tests it on synthetic data, and captures synthetic data affinity; the latter trains the model on synthetic data and tests it on real data, and reflects diversity. While subsequent work has analyzed TSTR from a Bayesian perspective~\cite{cas}, theoretical support for measuring affinity remains underdeveloped. Overall, these empirical metrics require comparisons within the same dataset, limiting their generalizability.
Another line of research assumes a \textit{hypersphere distribution} of data \cite{faithful} and calculates precision and recall of synthetic data relative to different support sets determined by the sphere radius. However, this hypersphere assumption and its dependence on a predefined distance function~\cite{regol2024categorical} restrict its tractability.

\nosection{Challenges in the Wireless Domain} Unlike image data, wireless data lacks semantic interpretability, precluding human visual validation. This necessitates theoretically grounded quality metrics to ensure reliability. Furthermore, wireless data exhibits diverse structural complexity, including spatial dependencies and dynamic channel variations. Since most wireless generative models are optimized for task- or dataset-specific requirements, fair quality assessment demands references for each model, in order to mitigate biases of data and task dependencies and maintain generalizability.

\section{Our Quality Metrics} \label{sec:metrics}
\begin{figure}[t]
    \centering
    \includegraphics[width=0.9\linewidth]{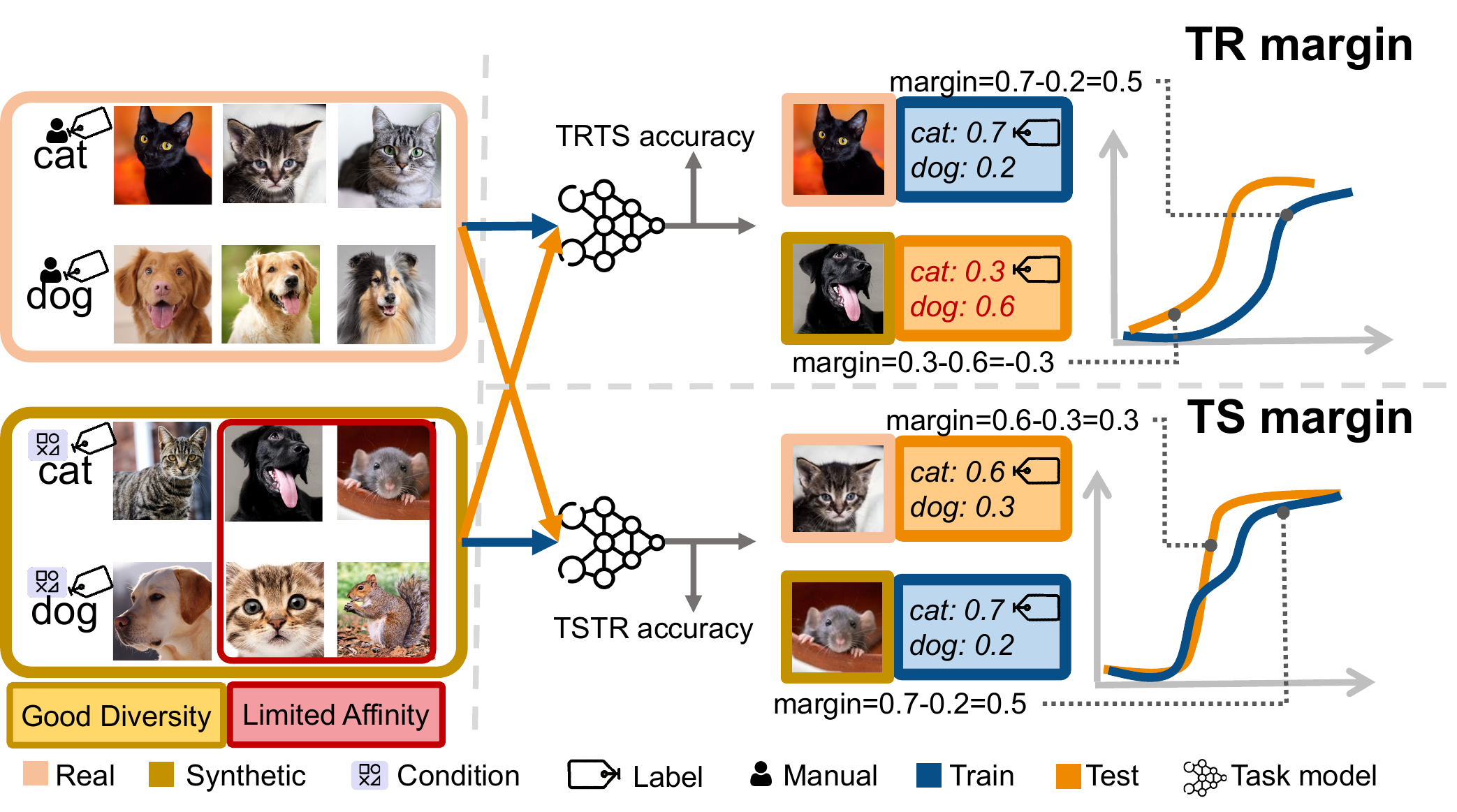}
    \vspace{-4mm}
    \caption{Our tractable and generalizable metrics: TR margin and TS margin. \textmd{Since wireless signals cannot be easily classified by humans, we use animal pictures to illustrate the basic ideas of confidence distribution and synthetic label errors.}}
    \vspace{-3mm}
    \label{fig:margin_cal}
\end{figure}

To address the limitations of existing evaluation metrics, we propose quality quantification metrics that balance tractability and generalizability. Specifically, our metrics inherit the idea of task-oriented evaluation with different train set choices, and complete Bayesian analysis~\cite{cas} to guarantee tractability. Furthermore, we repurpose the model decision confidence concept, known as \textit{margin}~\cite{aum}, to assess the quality of synthetic data, where the margin of the training set serves as a readily available reference. The quantification process is illustrated in \figref{fig:margin_cal}. In the following subsections, we provide a detailed explanation of the Bayesian framework and margin definition, along with empirical evidence demonstrating the advantages of our margin-based metrics.

\subsection{Metric Definition}
To establish the Bayesian theoretical foundation, we begin by formulating the two quality attributes, affinity and diversity, with conditional distributions between samples \( x \) and their corresponding labels \( y \), where the label of synthetic data corresponds to its generation condition. We use \( y \) to represent both labels and generation conditions interchangeably. Given the real data distribution \( p(x, y) \), a generative model approximates this distribution using its parameters \( \theta \), resulting in a synthetic data distribution \( p_{\theta}(x, y) \).

\begin{definition}[Affinity] Affinity measures similarity between synthetic samples and real samples, denoted by consistency between $p_{\theta}(x|y)$ and $p(x|y)$.
\end{definition}

\begin{definition}[Diversity] Diversity measures the extent that synthetic data covers the range of real data, denoted by consistency of $p_{\theta}(y|x)$ and $p(y|x)$.
\end{definition}

\nosection{Definition Explanation} Part (c) of \figref{fig:intro_fig1} illustrates the synthetic data distribution in terms of affinity and diversity. For high-affinity synthetic samples within one specific generation condition $y_0$, they resembles real samples of the same class $y_0$. Therefore, synthetic and real samples of class $y_0$ share similar distribution. In this way, affinity can be captured by similarity between $p_{\theta}(x|y)$ and $p(x|y)$. For high-diversity synthetic data, it should cover the range of real data across various generation conditions while maintaining the inter-class boundary. Such value coverage can be described as correlation between $p_{\theta}(y|x)$ and $p(y|x)$.
Based on the definition above, we further establish the connection between quality attributes task performance with different training set. 

\begin{theorem}[Affinity from train-real setup] When a task model is trained on real data, its test performance on synthetic data reflects affinity of synthetic data. High synthetic data affinity is a sufficient condition for good test performance in the train-real setup.
\end{theorem}

\begin{theorem}[Diversity from train-synthetic setup] When a task model is trained on synthetic data, its test performance on real data reflects the diversity of the synthetic data. High synthetic data diversity is a sufficient condition for achieving strong test performance in the train-synthetic setup\footnote{Proofs of the two theorems are included in Appendix \secref{theorem_proof}}.
\end{theorem}


A potential concerns emerges when synthetic data deviates from the real label space. While this issue is precluded by the affinity requirement of $p_{\theta}(x|y)=p(x|y)$, it remains unresolved by diversity requirement of $p_{\theta}(y|x)=p(y|x)$. By Bayes' theorem~\cite{bayes}, with known condition distribution $p(y)$, the diversity requirement implies \( \frac{p_\theta(x|y)}{p_\theta(x)} = \frac{p(x|y)}{p(x)} \), preserving the relative likelihoods of data conditioned on labels. As the ratio remains invariant despite the presence of label-mismatched samples, such synthetic data will maintain high test accuracy in the train-synthetic setup~\cite{cas}\footnote{ A generative model blending noise (with probability \(p\)) and real-distribution data (with probability \(1-p\)) still enables classifiers to generalize effectively to real data.}. Thus, conditional distribution alignment for diversity remains theoretically sound, and synthetic data remains effective for improving model performance provided that outlier samples are effectively identified and managed.


Having established the Bayesian foundation for using task performance as quality quantification metrics to ensure tractability, we further investigate task performance indicators to facilitate generalizable quality assessment. Previous metrics, such as TSTR and TRTS accuracies~\cite{good_gan}, can only compare synthetic data quality across multiple generative models on the same dataset, as the single-valued metric fails to indicate the absolute levels of affinity or diversity \textit{without a standard reference}. However, wireless generative models are tailored to specific hardware and signal properties, making unified comparisons across datasets challenging. To address this issue, we adopt a confidence distribution concept called margin to enable fair comparisons across datasets. The formal definition of margin along its generalization with standard reference is:


\begin{definition}[Margin] For a classification model, margin is the classification confidence difference between the label and the largest among other classes:
\begin{equation}
    \mathrm{Margin}(x,y)=z_y(x)-\max_{i\neq y} z_i(x)
\end{equation}
where $z(x)$ is the classification confidence of sample $x$. The margin calculation is demonstrated with concrete numerical and human-comprehensible examples in \figref{fig:margin_cal}.
\end{definition}

The margin distribution has been shown predictive of model generalization~\cite{bartlett2017spectrally,elsayed2018large}. 
For a well-generalized model, the margin distribution on test set closely resembles that of training set. Thus, the margin of training set serves as a \textit{natural standard reference}, enabling generalizable quality quantification: If the margin distribution of test set composed of synthetic data deviates significantly from that of real data training set, it indicates poor affinity of synthetic data. The same principle applies to the diversity property in train-synthetic setup.


\begin{conclusion}[Affinity with TR margin] The TR margin of test set, composed of synthetic data, reflects synthetic data affinity.
\end{conclusion}

\begin{conclusion}[Diversity with TS margin] The TS margin of test set, composed of real data, reflects synthetic data diversity.
\end{conclusion}

\subsection{Metric Advantage}
Given the fine granularity and natural reference points inherent in our quality metrics, we empirically investigate their correlation with task performance. To assess this relationship, we conduct controlled experiments by systematically varying the affinity and diversity of synthetic data, then evaluate model performance trained on real data and artificially modified synthetic data.

To examine affinity, we perturb synthetic data with Gaussian noise at varying standard deviations from 0.05 to 0.3 in increments of 0.05. The results reveal a strong positive correlation between the TR-margin metric and task performance (Pearson \(r = 0.900, p = 0.0146\)), outperforming the TRTS accuracy metric (\(r = 0.864, p = 0.0265\)). This suggests that margin-based metrics exhibits stronger correlation due to its fine-grained characterization of model behavior.
For diversity analysis, we vary the quantity of synthetic training data from 10\% to 100\% of the real dataset size in 10\% increments. Here, the TS-margin metric maintains a robust correlation with task performance (r = 0.864, p = 0.0013), while TSTR accuracy shows a weaker (r = 0.540) and statistically nonsignificant (p = 0.1147) relationship. These results demonstrate that margin-based metrics are more sensitive to variations in both affinity and diversity, making them more reliable for predicting task performance.

\section{Quality Assessment of Wireless Synthetic Data}

\begin{table*}[t]
    \caption{A taxonomy of existing data-driven wireless generative models is provided, with \colorbox{blue!10}{representative} works highlighted for further assessment in affinity and diversity attributes in this paper.}
    \vspace{-5pt}
    \centering
    \label{tab:literature}
    \begin{adjustbox}{max width=0.93\textwidth}

    \begin{tabular}{cccc}
    \toprule
    Literature &  Approach & Technique Category & Availability \\ 
    \midrule
   \colorbox{blue!10}{CsiGAN}~\cite{csigan}            & Synthetic data cross persons  & \multirow{3}{*}{\begin{tabular}[c]{@{}c@{}} Cross-domain transfer~\cite{nie2017medical}\end{tabular}}    & Open-source code and data  \\
    DFAR~\cite{wang2020cross}, LearnSense~\cite{LearnSense} & Synthetic data cross scenarios  & & Close-source \\
    SCTRFL~\cite{SCTRFL}, AutoFi~\cite{AutoFi} & Synthetic data cross time & & Close-source\\
    \hline
    \colorbox{blue!10}{RF-Diffusion} \cite{rf-diffusion}        & \multirow{2}{*}{\begin{tabular}[c]{@{}c@{}}Training set enrichment \\ with synthetic samples\end{tabular}}    & \multirow{2}{*}{\begin{tabular}[c]{@{}c@{}}In-domain  amplification~\cite{gan16}\end{tabular}}  & Open-source code    \\
    RF-EATS~\cite{rf-eats}, AF-DCGAN~\cite{AF-DCGAN}           &   & & Close-source \\
    \bottomrule
    \end{tabular}
    \end{adjustbox}
    \vspace{-5pt}
\end{table*}
To systematically assess the quality of wireless synthetic data, we conduct a comprehensive literature review of wireless generative models, aligning them to corresponding techniques in the AI research and categorizing them into two distinct groups. We re-implement representative works from each category using a unified experimental framework to ensure a fair comparison, and evaluate the affinity and diversity of synthetic data from each model.

\subsection{Wireless Generative Model Taxonomy}
\label{sec:literature}
\begin{figure}[t]
    \centering
    \includegraphics[width=.7\linewidth]{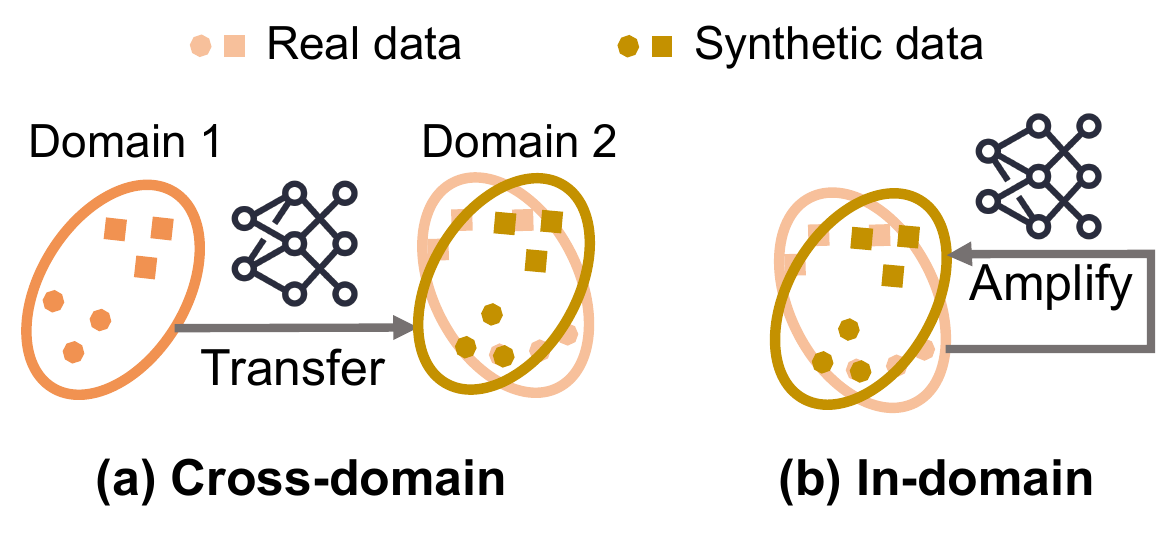}
    \vspace{-5mm}
    \caption{Categories of generative model techniques.}
    \vspace{-5mm}
    \label{fig:generative_category}
\end{figure}

We review and categorize recent publications that augment training sets with synthetic data in \tabref{tab:literature}. They are categorized based on source and target data domains for synthesis in \figref{fig:generative_category}:

\nosection{Cross-domain Sample Transfer}
This category uses data from one domain as input and produce synthetic data in a different domain. Here, a `domain' refers to application-specific aspects such as individuals, environments, or device configurations. The goal is to transfer real samples from source domain into synthetic samples in the target domain and complement the target domain data.

\nosection{In-domain Sample Amplification}
This category focuses on generating additional samples within the same domain. While similar to data augmentation, this approach replaces heuristic rules with data-driven generative models to create new synthetic samples and enhance dataset diversity.



\subsection{Quality Quantification and Analysis}
\label{ssec:empirical_quality}
Based on our quantification metrics and categories of existing wireless generative models, we conduct a series of experiments to answer the following research question:

\begin{tcolorbox}[size=small]
    \textbf{RQ1:} In terms of affinity and diversity, how is the quality of existing wireless synthetic data?
\end{tcolorbox}



\nosection{Representative Generative Models} 
We aim to benchmark all existing wireless generative models shown in \figref{fig:generative_category}. However, the accessibility of both models and datasets varies significantly, which influenced our selection process. To ensure a fair and unbiased evaluation, we prioritize recent models that provide open-source code and publicly available dataset partitions, as this helps mitigate implementation biases and enhances the reliability of our results.

For cross-domain transfer, we select CsiGAN~\cite{csigan}, which transfers gesture samples from trained users to unseen users. The training data consists of only a subset of gesture classes, while the testing data includes all classes. This setup, with incomplete class coverage during training, reflects real-world deployment scenarios where samples from new users are often limited. For in-domain amplification, we select RF-Diffusion~\cite{rf-diffusion}, a diffusion model conditioned on human orientation, activity, and antenna configuration to generate synthetic samples.  It utilizes all available activity classes for training to ensure comprehensive coverage of the conditions.

\nosection{Task model and dataset}
We employ a simple yet effective ResNet34 architecture~\cite{resnet} as the task model, which is proven effective across various wireless tasks and benchmarks~\cite{yang2023benchmark}. The selected generative models are tested on the datasets used in their original studies: the SignFi dataset~\cite{signfi} for CsiGAN\cite{csigan}, and the Widar dataset~\cite{widar} for RF-Diffusion~\cite{rf-diffusion}. 

\begin{figure}[t]
    \centering
    \includegraphics[width=.75\linewidth]{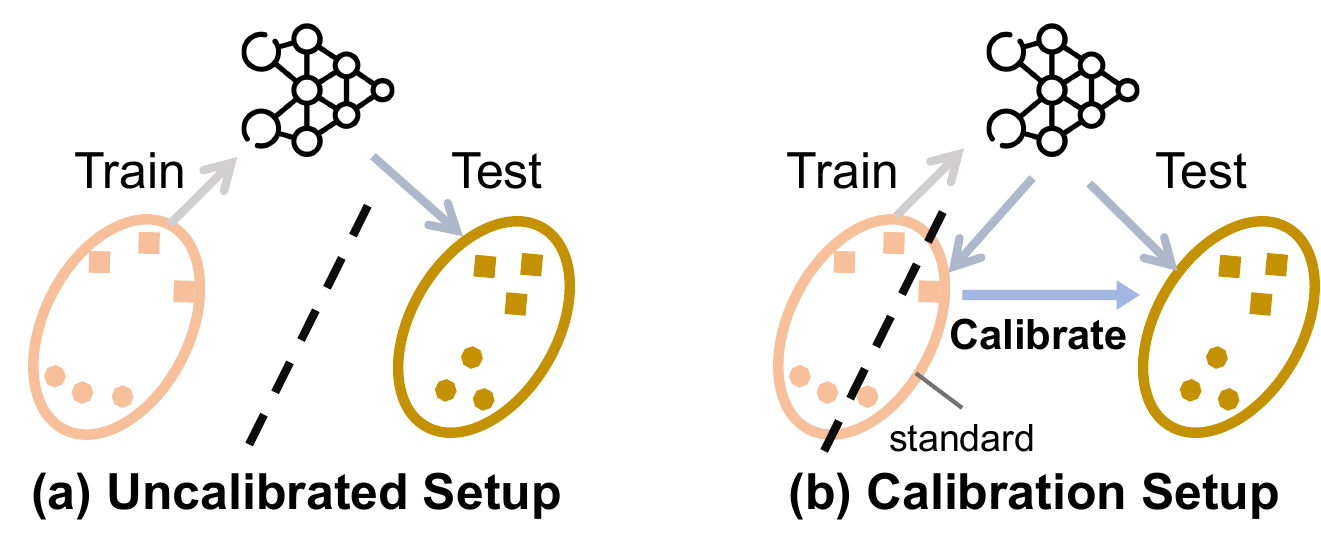}
    \vspace{-5mm}
    \caption{Calibration of the margin.}
    \vspace{-5mm}
    \label{fig:calibration}
\end{figure}

\nosection{Margin Calibration}
Our metrics employ margin comparisons between the test set and the train set to enable generalizable quality quantification across datasets, where the margin of train set serves as a reference for each dataset. However, inherent performance discrepancies between the train and test sets~\cite{keskar2016large, novak2018sensitivity} can disrupt the margin comparison. To address this issue, we introduce a calibration procedure shown in \figref{fig:calibration}. Specifically, in both the TR and TS setups, we pre-select a subset of the corresponding training data and designate it as the standard test set, ensuring it follows the same distribution as the train set. The calibration process involves calculating the mean margin values for both the train and standard test sets, and then adjusting the margins of the actual test set based on the difference between these two mean margin values. The data quality is quantified by measuring the distribution gap between the training set and calibrated test set. One suitable measure for this gap is Jensen-Shannon (JS) Divergence~\cite{menendez1997jensen}, which assesses the discrepancy between two distributions, and a smaller value indicates greater similarity. The results of quality assessment with JS Divergence, are presented in \tabref{tab:metric_cmp}.

\nosection{Necessity of Calibration} To validate the importance of calibration, we evaluate TR-margin and TS-margin across different model architectures (ResNet18, ResNet34) and training epochs, involving model accuracy variations exceeding 20\%. The analysis reveals that uncalibrated margin distribution gap show 13.5\% variations and calibration achieves better consistency of 5.9\% variation. These findings demonstrate that calibration mitigates model dependency to some extent, enabling more stable and reliable quality assessment.



\nosection{Data Quality Results}
The synthetic data quality results are shown in \figref{fig:metrics}: A stronger correlation between the margins of the train set and test set in the train-real and train-synthetic setups indicates better affinity and diversity, respectively. Based on these quality quantification results, we highlight the following key observations:


\begin{figure}[t]
	\centering
    \tightsubfigures
    \begin{subfigure}{.49\linewidth}
	\includegraphics[width=\linewidth]{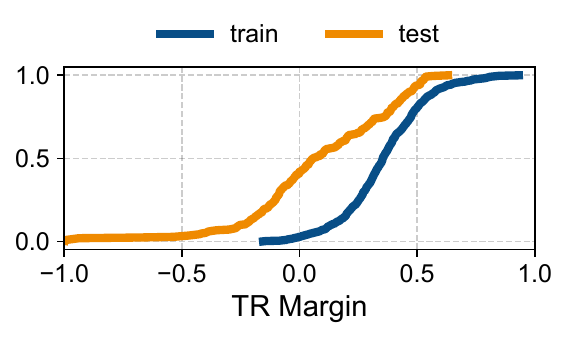}
		\caption{Affinity of Cross-domain}
		\label{fig:affinity_cross}
	\end{subfigure}
    \begin{subfigure}{.49\linewidth}
	\includegraphics[width=\linewidth]{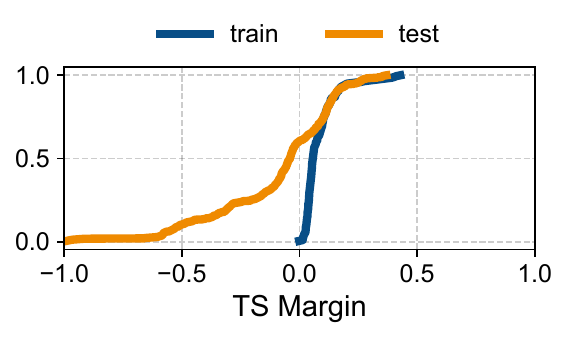}
		\caption{Diversity of Cross-domain}
		\label{fig:diversity_cross}
	\end{subfigure}
    \begin{subfigure}{.49\linewidth}
	\includegraphics[width=\linewidth]{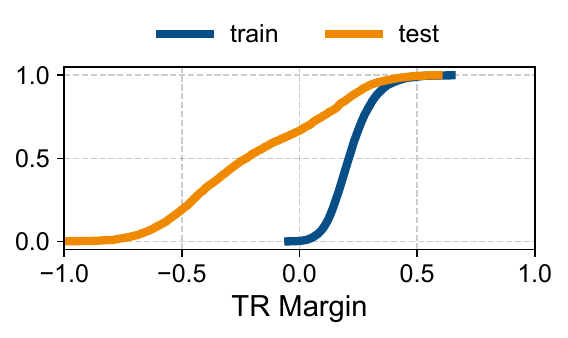}
		\caption{Affinity of In-domain}
		\label{fig:affinity_in}
	\end{subfigure}
    \begin{subfigure}{.49\linewidth}
	\includegraphics[width=\linewidth]{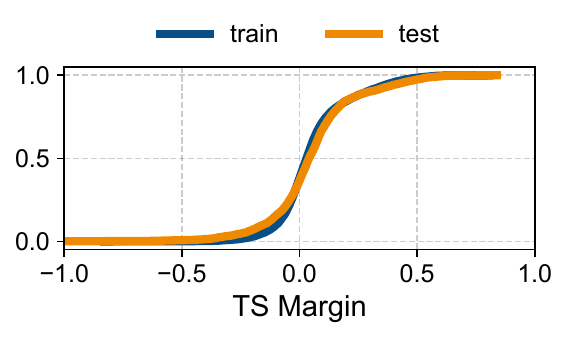}
		\caption{Diversity of In-domain}
		\label{fig:diversity_in}
	\end{subfigure}
    \vspace{-3mm}
	\caption{Quality quantification of synthetic data.}
    \vspace{-5mm}
	\label{fig:metrics}
\end{figure}

\begin{itemize}
    \item \textit{Prevalent Affinity Limitation.} In both cross-domain (\figref{fig:affinity_cross}) and in-domain (\figref{fig:affinity_in}) generation, the calibrated test margins for synthetic data show significant discrepancies from the train margin. This margin difference indicates that the synthetic data is \textit{affinity-deficient and mislabeled}, which can mislead the task model's training.
    \item \textit{Imbalanced Diversity in Cross-domain Generation.} In the TS margin (\figref{fig:diversity_cross}) for cross-domain generation, the calibrated test margin closely aligns with the train margin for the top 40\% of synthetic samples, but diverges significantly in the remaining 60\%. This discrepancy indicates inadequate and imbalanced diversity in the cross-domain synthetic data.
    \item \textit{Good Diversity in In-domain Generation.} The TS test margin (\figref{fig:diversity_in}) for in-domain generation closely matches the train margin, indicating good diversity of in-domain synthetic data.
    \item \textit{Abnormal Train Margins.} An additional observation in the train-synthetic setup is the abnormal distribution of train margins for in-domain generation. In other setups, train margins are predominantly greater than zero, which is reasonable for a model well-behaved on the training set. However, in \figref{fig:diversity_in} approximately 20\% of the train margins are negative. This abnormal distribution suggests label confusion in the in-domain synthetic data, which disrupts the model's training process.
\end{itemize}

\begin{tcolorbox}[size=small]
\textbf{Answer to RQ1}: Our quantification results reveal that existing wireless synthetic data exhibits prevalent limited affinity, with imbalanced diversity for cross-domain generation and good diversity for in-domain generation.
\end{tcolorbox}

\subsection{Investigation of Underlying Causes}\label{ssec:weakness}
We investigate the underlying causes of the prevalent low affinity and partially imbalanced diversity in wireless synthetic data and answer the following research question:
\begin{tcolorbox}[size=small] 
\textbf{RQ2:} What are the causes of low affinity and imbalanced diversity in the wireless synthetic data? 
\end{tcolorbox}

\nosection{Analysis Method}
We analyze the potential causes of synthetic data quality issues throughout the entire generation process, including pre-generation preparation and domain-specific post-generation processing. Based on guidelines above, we identify limitations of the representative generative models, which further inspire the development of our quality-aware synthetic data utilization scheme.

\begin{figure}[t]
	\centering
    \begin{subfigure}{.49\linewidth}
	\includegraphics[width=\linewidth]{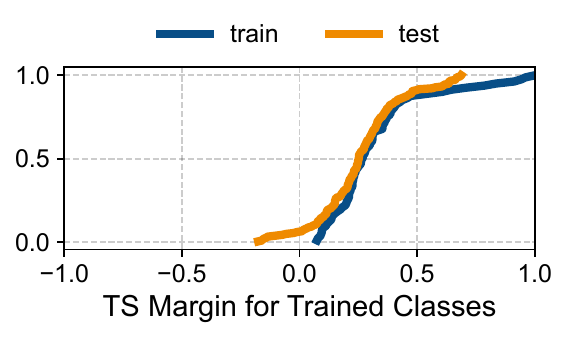}
	\end{subfigure}
    \begin{subfigure}{.49\linewidth}
	\includegraphics[width=\linewidth]{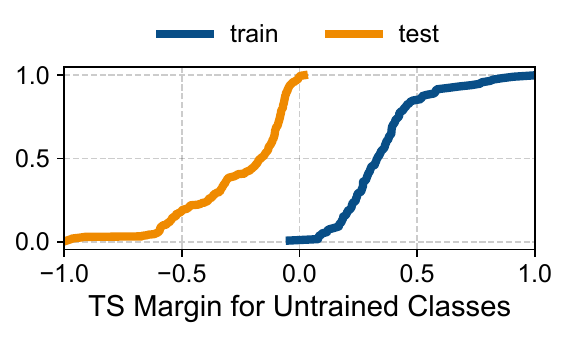}
	\end{subfigure}
    \vspace{-4mm}
	\caption{Diversity comparison between trained and untrained classes of generative models.}
    \vspace{-4mm}
	\label{fig:trained_untrained}
\end{figure}

\nosection{Incapability of Condition Transfer}
The pre-generation phase prepares training data for generative models, where models designed for cross-domain transfer often encounter incomplete class coverage in the training set~\cite{csigan, wiag, wang2020cross}. This issue arises from the specific constraints of wireless applications, which aim to augment the target domain with limited available data. To further investigate the effect of class coverage, we compare the affinity and diversity between trained and untrained classes of generative models in \figref{fig:trained_untrained}. The TS margins for the train and test sets of the trained classes are well aligned, while the margins for the untrained classes are more divergent. This margin distribution difference shows a significant diversity advantage of trained classes over untrained ones. The diversity gap between trained and untrained classes underscores the challenges of effective cross-domain condition transfer.

\begin{figure}[t]
    \centering
    \includegraphics[width=0.9\linewidth]{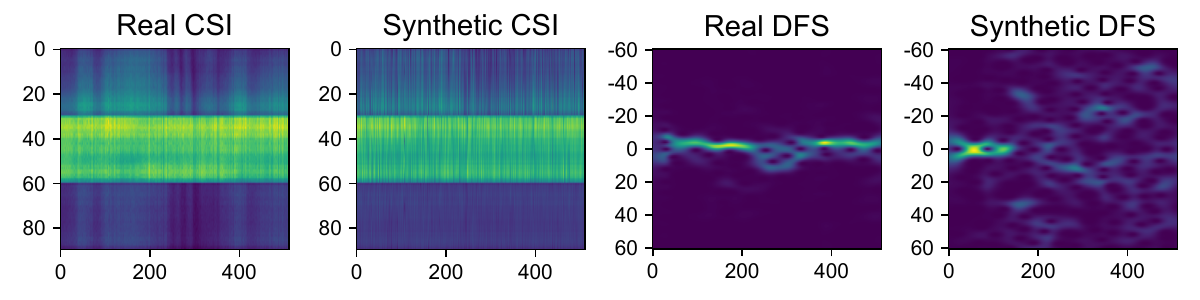}
    \vspace{-5mm}
    \caption{Data with Similar CSIs but Dissimilar DFSs.}
    \vspace{-5mm} 
    \label{fig:csi_dfs}
\end{figure}

\nosection{Unawareness of Domain-specific Processing} 
The post-generation phase transforms synthetic data into application-specific formats. A typical example is WiFi signal processing, where generative models synthesize raw channel state information (CSI)~\cite{rf-diffusion,diffar}, and applications like human action recognition rely on higher-level features such as Doppler frequency shift (DFS)~\cite{rfboost} and Body Velocity Profile (BVP)~\cite{widar}. The discrepancy in signal formats arises from the differing objectives of generative models and applications: CSI synthesis captures diverse environmental factors and is effective for complementing real data in complex scenarios, whereas DFS and BVP are designed to be resilient to such variations and focus on target tasks. This divergence between the model's raw output (CSI) and application's input (DFS) is overlooked in prior work. Our analysis of RF-Diffusion highlights this issue and reveals that generative models trained on CSI fail to generate DFS similar to the real DFS, as shown in \figref{fig:csi_dfs}. In summary, generative models that do not account for domain-specific processing fail to align their outputs with the particular semantic properties required by target applications.

\nosection{Insights for Better Utilization}
The entire generation process of wireless synthetic data is inherently tied to domain-specific properties. While we have identified quality issues at both pre-generation and post-generation stages, the variety of generation techniques and applications makes this investigation far from exhaustive.  Given the difficulty of comprehensive quality assurance throughout data generation, an effective utilization scheme for synthetic data should incorporate domain-specific knowledge, and overcome the quality limitations in a \textit{task-oriented} and \textit{generic} manner. To this end, both our quality metrics and utilization approach leverage task models to analyze synthetic data. 
Since affinity deficiency directly impacts label correctness and compromises training data, and diversity imbalance results in uneven performance gains across classes, our first step in developing a better utilization scheme addresses the affinity limitation.



\begin{tcolorbox}[size=small]
\textbf{Answer to RQ2}: Generative models agnostic to untrained condition distributions and domain-specific processing produce low-quality data. An effective utilization scheme for synthetic data should recognize the quality issues and underlying causes. 
\end{tcolorbox}

\section{Quality-aware Utilization of Synthetic Data}

\begin{figure*}[t]
    \centering
    \includegraphics[width=0.85\linewidth]{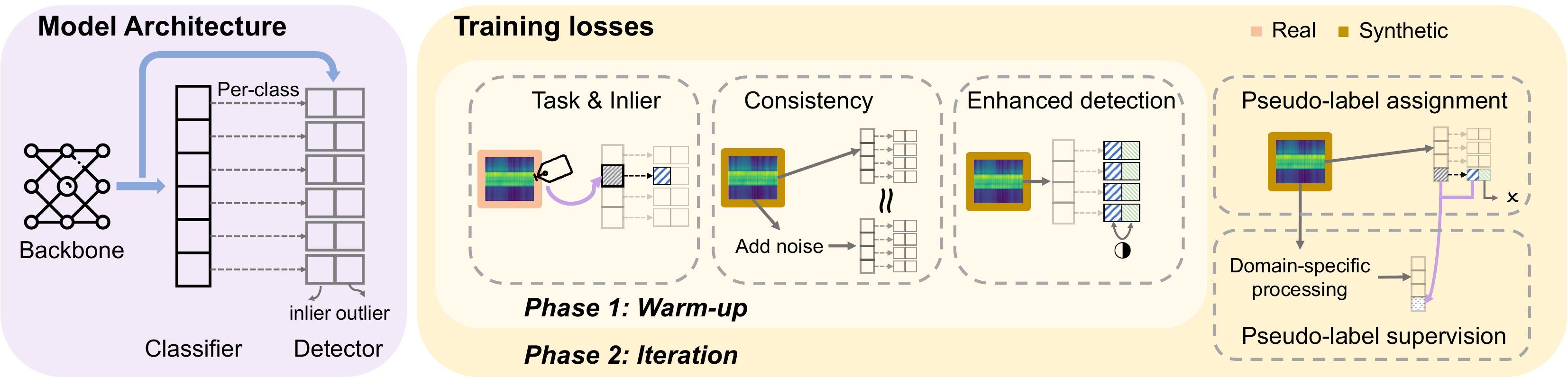}
    \vspace{-2mm}
    \caption{\sys Workflow.}
    \vspace{-5mm}
    \label{fig:system workflow}
\end{figure*}

Building on the quality quantification of affinity and diversity in synthetic data (\textbf{RQ1}), and exploring the causes of quality issues (\textbf{RQ2}), we conclude that the primary limitation of current wireless synthetic data is prevalent affinity deficiency, which leads to mislabeled synthetic data and degraded task performance. Keeping these observations in mind, we address the following research question:

\begin{tcolorbox}[size=small] 
    \textbf{RQ3:} Given synthetic data with limited affinity, how can it be utilized to improve task performance in a generic way? 
\end{tcolorbox}


\subsection{Approach Overview}
We propose a quality-guided utilization scheme for synthetic data in a post-processing framework, adaptable to various generative models. This approach stems from the cause investigation that data quality is deeply intertwined with the entire generation process, making it challenging to design targeted solutions to cover every aspect of quality issues. 

In contrast to previous methods that rely on condition-labeled synthetic data for fully supervised training~\cite{csigan, rf-diffusion}, our solution, \sys, eliminates the need for generation conditions as labels. Instead, it adopts a semi-supervised learning paradigm, leveraging labeled real data and unlabeled synthetic data. The model simultaneously performs the target task while assigning pseudo-labels to the synthetic samples. The pseudo-labels provide additional supervision with domain-specific processing to iteratively refine the model.
Furthermore, since the actual label of a synthetic sample may not correspond to any real sample, we adopt an \textit{open-set} setup~\cite{park2022opencos, li2023iomatch}, as opposed to the \textit{closed-set} setup. The closed-set setup assumes that the synthetic data classes are identical to those of the real data. In contrast, the open-set setup assumes that synthetic data may contain classes not present in real data. The samples falling outside the real data classes is identified as irrelevant and filtered out.


\nosection{\sys Model Architecture} The \sys model comprises three main components: backbone, task classifier and inlier-outlier detector, as shown in \figref{fig:system workflow}. Specifically, the backbone model serves as a feature extractor for both classification and inlier-outlier detection. The classifier, denoted as $f$, is responsible for the target classification task and its output dimension is the number of classes $C$ in real data. The detector decides whether the input sample is an inlier or outlier of a specific class. With the detector $g^j$ for class $j$, the probability of a sample $x$ being an inlier is $g^j(t=0|x_b)$, while the probability of $x$ being an outlier is $g^j(t=1|x_b)$. This \textit{one-vs-all} detection~\cite{saito2021ovanet} has proven effective in identifying per-class inliers. 

\nosection{\sys Workflow} The semi-supervised learning of \sys comprises two phases: model warm-up with labeled real data and unlabeled synthetic data, and model iteration with labeled real data and pseudo-labeled synthetic data. The training losses for each phase are shown in \figref{fig:system workflow}. In the first phase, the task classification and inlier-outlier detection models are trained in a supervised manner, followed by an unsupervised enhancement to further improve the robustness of both classification and detection. With a well-behaved model on classification and detection, the second phase introduces additional training losses by assigning pseudo-labels to the filtered synthetic data. These pseudo-labels, combined with application-specific processing, provide additional supervision that incorporates domain knowledge to guide model training.

\subsection{Phase 1: Model Warm-up}
The first phase focuses on training a model to achieve good performance in both task classification and inlier-outlier detection, preparing it for effective utilization of synthetic data in the subsequent phase. The model warm-up phase involves two key components: supervised training using labeled real data and self-supervised training with unlabeled synthetic data. 

\nosection{Supervision with Labeled Real Data}
Using manually collected real data $x$ with reliable labels $y$, we train the target task classifier $f$ following common practice. Besides, we enforce the labeled real samples to be inliers of their label classes, and outliers of all other classes, as determined by the one-vs-all detectors $g^j$s. The supervised training losses with labeled real data are as follows:
\begin{equation}
    \begin{split}
        L_s &= L_{\text{task}} + L_{\text{ova}} \\ 
        L_{\text{task}} &= \mathrm{cross\ entropy}(f(x),y) \\
        L_{\text{ova}} &= -\log(g^{y}(t=1|x) - \min\nolimits_{j\neq y} \log(g^j(t=0|x))
    \end{split}
\end{equation}

\nosection{Self-supervision with Unlabeled Synthetic Data}
With the low-affinity synthetic data $u$ without usable labels, we employ consistency regularization to ensure robust model predictions and entropy minimization to enhance the distinction between inliers and outliers. The rationale behind consistency regularization is based on the idea that for an input sample with small perturbations, the model's prediction should remain similar to that of the original unperturbed sample~\cite{laine2016temporal}, as long as the noise does not interfere with the task-specific semantics. For the wireless signals, we use Gaussian noise, a widely recognized model for propagation channel noise~\cite{proakis2008digital}. Furthermore, to encourage clearer discrimination between class inliers and outliers, we minimize the entropy of the outputs from the one-vs-all detectors $g^j$ to avoid ambiguous inlier-outlier decision. The training losses for enhancing consistency and distinction with unlabeled synthetic data include:
\begin{equation}
\resizebox{0.9\linewidth}{!}{$
    \begin{split}
        L_{u} &= L_{\text{cons}} + L_{\text{ent}} \\
        L_{\text{cons}} &= \sum\nolimits_j\lVert g^j(u) - g^j(u+\mathrm{noise}) \rVert_2^2 + \lVert f(u) -f(u+\mathrm{noise})\rVert \\
        L_{\text{ent}} &= \sum\nolimits_j\sum\nolimits_{t=0}^1 -g^j(t|u)\log(g^j(t|u))
    \end{split}
$}
\end{equation}

\subsection{Phase 2: Model Iteration}
After the warm-up phase, the model is now capable of predicting input classes and distinguishing between inliers and outliers. In this phase, we use the model to filter out synthetic data with low affinity (outliers) and assign pseudo-labels to the remaining high-affinity synthetic data (inliers). The pseudo-label supervision iteratively improves the model's performance in a domain-dependent manner.

\nosection{Pseudo-label Assignment}
For each synthetic sample $u$, we first determine its most likely class $\hat{y}$ using the task classifier $f$. We then assess its affinity by passing it through the inlier-outlier detector $g^{\hat{y}}$. If the sample is classified as an inlier of class $\hat{y}$, it is assigned the pseudo-label $\hat{y}$ and can provide further supervision for the task.

\nosection{Pseudo-label Supervision with Domain Knowledge} So far a subset of synthetic samples (denoted as $\mathcal{S}$) has been identified as inliers and assigned pseudo-labels. To further improve the model, we employ a pseudo-label utilization strategy in which the model's predictions for augmented samples should align with the assigned pseudo-labels~\cite{weak_strong}. What sets \sys apart is the incorporation of wireless-specific augmentations~\cite{rfboost, aug2, aug3}, such as varying time window lengths and random subcarrier distortion, to integrate domain-specific knowledge into the label prediction process.
The training loss for inlier synthetic samples $s_b\in \mathcal{S}$ with application-specific augmentation $\mathcal{A}_s$ is
\begin{equation}
    L_{\text{pseu}} = \mathrm{cross\ entropy}(\mathcal{A}_s(s_b),\hat{y})
\end{equation}

\begin{tcolorbox}[size=small]
\textbf{Answer to RQ3}: We propose a semi-supervised learning framework that treats synthetic data as unlabeled and selects high-affinity synthetic samples to aid model training. 
\end{tcolorbox}

\section{Implementation}
We reproduce existing generative models and implement our semi-supervised learning with PyTorch~\cite{pytorch}. For CsiGAN, we attempt to align our results with the original paper, as the authors have publicly released their dataset processing scripts~\cite{csigan_code}. For RF-Diffusion, since the dataset processing code is not open-sourced, we use the public dataset partition from RFBoost~\cite{rfboost_code} for the Widar dataset~\cite{widar}. The model backbone is primarily ResNet34~\cite{resnet}, while we also include microbenchmarks to investigate the effect of backbones.

Model training and inference were performed on a workstation equipped with a Genuine Intel (R) CPU and six NVIDIA 3090 GPUs. For the evaluation of CsiGAN, the models were trained for 100 epochs with a learning rate of \(2 \times 10^{-4}\), a batch size of 16, and a StepLR scheduler configured to decay the learning rate by a factor of 0.1 every 40 epochs. For the evaluation of RF-Diffusion, the models were trained for 10 epochs with a learning rate of \(1 \times 10^{-3}\), a batch size of 64, and a StepLR scheduler set to decay the learning rate by a factor of 0.9 every 2 epochs.  These configurations were based on the size and convergence characteristics of each dataset.


\section{Evaluation}
\subsection{Experiment Setup}
\subsubsection{Dataset and Generative Models}

Our empirical analysis focuses on two representative generative models: CsiGAN for cross-domain transfer and RF-Diffusion for in-domain amplification. Unless otherwise specified, the initial amount of synthetic data is the same as real data of the original dataset in our experiments.


\nosection{Cross-domain Transfer}
CsiGAN is trained on the SignFi dataset~\cite{signfi}, which contains WiFi CSI traces of 5 users performing 150 actions. Each sample consists of 200 packets captured with 3 receiving antennas and 30 subcarriers, resulting in a shape of $200 \times 30 \times 3$. Following the original experimental design, we treat each user as an independent domain due to their unique behavioral characteristics. For model training, we utilize a subset of 1,000 real samples, consistent with the original implementation.

\nosection{In-domain Amplification}
RF-Diffusion is trained on the Widar dataset~\cite{widar}, a widely used dataset for gesture recognition with WiFi. This dataset includes CSI traces from 17 users performing 22 gestures in 5 different locations and orientations. Each sample from a single receiver has the shape of $T\times 90$, where $T$ is the packet amount. As its data partition is not open sourced, we employ the same dataset partition of RFBoost~\cite{rfboost} with the 6 most frequent gestures from 15 users across three distinct scenarios, whose train set has 13,275 samples.

\subsubsection{Evaluation Metrics} 
Given the goal of task performance enhancement with synthetic data, 
we use task performance as the primary evaluation criterion. We also re-assess the affinity and diversity of filtered synthetic data. The enhanced task performance and refined data quality will underscore the necessity of quality-guided utilization and the effectiveness of \sys in selecting high-quality synthetic data.


\subsubsection{Baseline Methods for Comparison} 
We compare the performance of \sys against the following baselines: (1) \textit{Real data only}, representing traditional training without synthetic data; (2) \textit{Nonselective mixture} of synthetic and real data, a common practice in prior work (see \tabref{tab:literature}); (3) Synthetic data \textit{filtering with SSIM}, which selects synthetic data based on visual similarity to real data; and (4) Synthetic data \textit{filtering with TRTS}, which trains a model on real data and excludes synthetic samples whose model predictions deviate from their labels (generation conditions). In contrast, \sys introduces iterative filtering and pseudo-label assignment for synthetic data utilization.
Our evaluation aims to answer the following questions: 
\begin{tcolorbox}[size=small]
    \textbf{RQ4:} How does \sys perform in task performance and data quality attributes of affinity and diversity?
    
    \textbf{RQ5:} What is the impact of design choices, including loss components, backbone model, synthetic data volume, and label origin, in the \sys framework?

    \textbf{RQ6:} What is the role of conditions in synthetic data generation and how do they affect data quality?

\end{tcolorbox}


\subsection{Overall Performance}
\label{ssec:overall_performance}

\begin{figure}[t]
    \centering
    \includegraphics[width=0.92\linewidth]{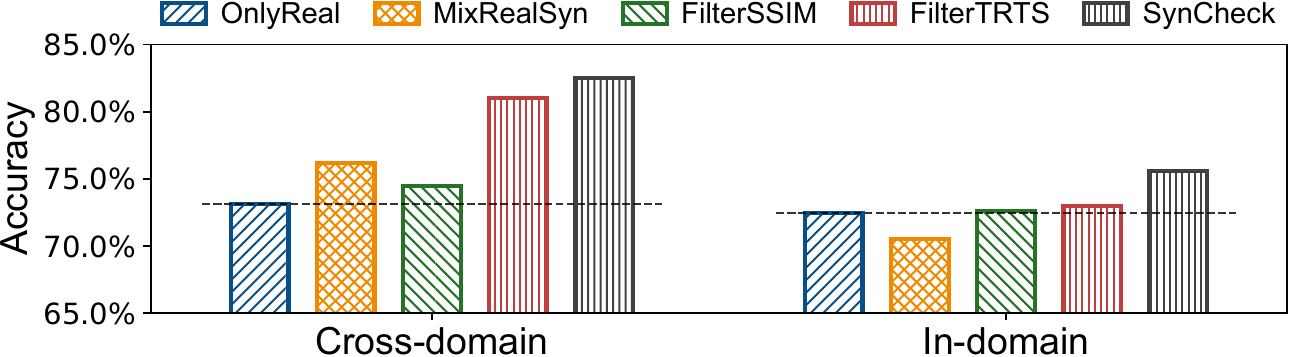}
    \vspace{-3mm}
    \caption{Overall Performance.}
    \vspace{-3mm}
    \label{fig:overall_performance}
\end{figure}

\nosection{End-to-end Performance on Target Task}
The performance of \sys on the target task, compared to baseline methods, is shown in \figref{fig:overall_performance}. The dashed line represents training with only real data. \sys outperforms all baselines, achieving 82.7\% in cross-domain transfer and 75.6\% in in-domain amplification, surpassing the nonselective mixture of synthetic and real data (76.2\% and 70.5\%) and the real-data-only baseline (73.2\% and 72.5\%). This represents an 8.6\% and 7.2\% improvement over nonselective utilization, highlighting \sys's effectiveness in quality-guided synthetic data utilization. Performance degradation in the nonselective mixture stems from affinity limitations in cross-domain scenarios, while diversity gains partially mitigate this in in-domain setups.
Synthetic data filtering with SSIM achieves 74.5\% and 72.6\%, showing minor gains over real data by addressing affinity through visual similarity. A stronger baseline, TRTS filtering, achieves 81.0\% and 73.0\%, demonstrating the advantage of task-oriented selection. However, TRTS filtering still falls short of \sys, as it relies on generation conditions as labels and causes mislabeled data.




\begin{figure}[t]
	\centering
    \begin{subfigure}{.49\linewidth}
	\includegraphics[width=\linewidth]{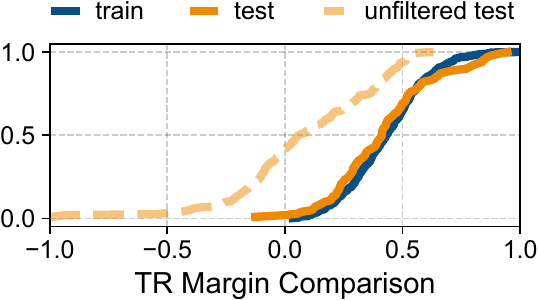}
        \vspace{-5mm}
		\caption{Affinity of Cross-domain}
		\label{fig:after_affinity_cross}
        \vspace{2mm}
	\end{subfigure}
    \begin{subfigure}{.49\linewidth}
	\includegraphics[width=\linewidth]{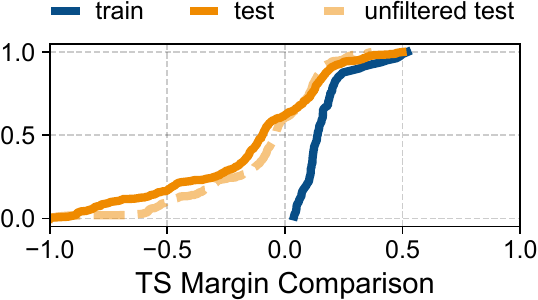}
        \vspace{-5mm}
		\caption{Diversity of Cross-domain}
		\label{fig:after_diversity_cross}
        \vspace{2mm}
	\end{subfigure}
    \begin{subfigure}{.49\linewidth}
	\includegraphics[width=\linewidth]{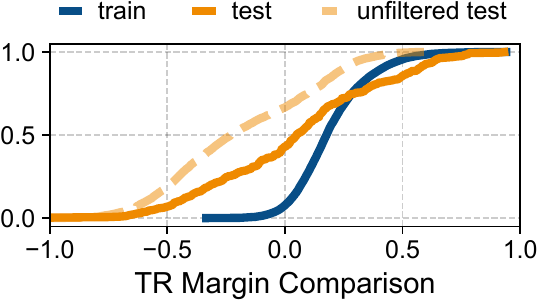}
        \vspace{-5mm}
		\caption{Affinity of In-Domain}
		\label{fig:after_affinity_in}
	\end{subfigure}
    \begin{subfigure}{.49\linewidth}
	\includegraphics[width=\linewidth]{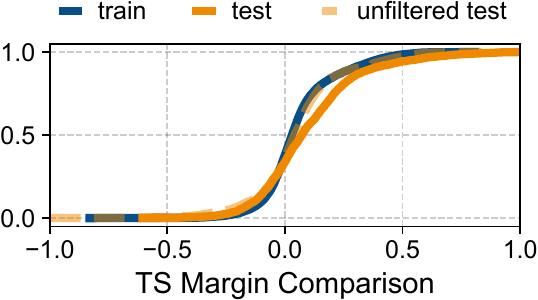}
        \vspace{-5mm}
		\caption{Diversity of In-Domain}
		\label{fig:after_diversity_in}
	\end{subfigure}
    \vspace{-3mm}
	\caption{Quality re-quantification of the filtered synthetic data with \sys compared to original unfiltered data.}
	\label{fig:reeval_metrics}
    \vspace{-1mm}
\end{figure}

\nosection{Quality Re-evaluation}
We re-evaluate the quality attributes of affinity and diversity for the filtered and relabeled synthetic data with \sys, as shown by the `test' in \figref{fig:reeval_metrics}. In contrast, the quality assessment of unfiltered synthetic data is denoted as `unfiltered test'. The filtered synthetic data demonstrates improved correlation between the TR margins of the train and test sets in both cross-domain and in-domain generation, indicating \textit{improved affinity}. Moreover, the distribution of the TS margin before and after filtering remains similar, suggesting \textit{preserved diversity}.
To quantify this improvement, we calculate the JS Divergence between the margins of the training set and test set in \tabref{tab:metric_cmp}.  Our results show that \sys significantly improves the TR-margin, achieving a 57.3\% and 29.5\% reduction in JS Divergence for TR margins, thereby enhancing data affinity. In contrast, the TS margin divergence increases only slightly, preserving the diversity of synthetic data.
This quality enhancement aligns with task performance, further validating the effectiveness of our quality assessment metrics. 


\begin{tcolorbox}[size=small]
\textbf{Answer to RQ4}: \sys achieves superior task performance and enhanced data affinity. These results underscore the importance of quality-guided utilization of synthetic data.

\end{tcolorbox}

\begin{table}[t]
    \caption{Comparison of affinity and diversity between the original and \sys filtered synthetic data, where smaller scores indicates better affinity and diversity, respectively.}
    \vspace{-3mm}
    \centering
    \label{tab:metric_cmp}
    \begin{adjustbox}{max width=0.9\linewidth}
    \begin{tabular}{ccccc}
    \toprule
    & \multicolumn{2}{c}{\textbf{JS Div of TR Margin} ($\downarrow$)} & \multicolumn{2}{c}{\textbf{JS Div of TS Margin} ($\downarrow$)} \\
    Methods & Unfiltered & \sys & Unfiltered & \sys \\
    \midrule
    CsiGAN~\cite{csigan}    & 0.503 & 0.215 & 0.640 &  0.657   \\
    RF-Diffusion~\cite{rf-diffusion}  & 0.694 & 0.489 & 0.135 &   0.176      \\
    \bottomrule
    \end{tabular}
    \end{adjustbox}
    \vspace{-4mm}
\end{table}

\subsection{Ablation Studies}
We conduct several ablation studies to validate the effectiveness and necessity of the loss function components in the semi-supervised learning framework. We categorize these losses into two groups based on their functions:
(1) The one-vs-all loss \( L_{ova} \) for per-class inlier-outlier detectors \( g^j \), which are trained on real labeled data to facilitate outlier detection on unlabeled synthetic data. 
(2) The consistency loss \( L_{cons} \) and entropy loss \( L_{ent} \), applied to unlabeled synthetic data. These losses are designed to enforce the prediction robustness and discrimination of synthetic data for the target task and inlier-outlier detection.  

\begin{table}[!htbp]
    \caption{Ablation Studies for Loss Function Design.}
    \vspace{-4mm}
    \centering
    \label{tab:ablation_study}
    \begin{adjustbox}{max width=0.9\linewidth}
    \begin{tabular}{ccc}
    \toprule
    Design Choices &  Cross-domain Acc. & In-domain Acc. \\ 
    \midrule
    Full \sys            & \textbf{0.827} &  \textbf{0.756}     \\
    w/o $L_{\text{ova}}$ & 0.760 & 0.746 \\
    w/o $L_{\text{cons}}$ and $L_{\text{ent}}$ & 0.788 & 0.751 \\
    \midrule
    Nonselective mixture & 0.762 & 0.705 \\
    \bottomrule
    \end{tabular}
    \end{adjustbox}
    \vspace{-3mm}
\end{table}

\tabref{tab:ablation_study} presents the 
task performance comparison for ablated models. The original \sys achieves the highest performance in both in-domain and cross-domain setups, underscoring the importance of all loss components. In contrast, removing the one-vs-all loss \( L_{\text{ova}} \) leads to a 8.10\% and 1.3\% drop in task accuracy for cross-domain and in-domain setups, respectively. Similarly, removing the consistency and entropy constraints (\( L_{\text{cons}} \) and \( L_{\text{ent}} \)) on unlabeled synthetic data results in a 4.72\% and 0.67\% decline in performance. 

\subsection{Microbenchmark}
We analyze the impact of the backbone models, original synthetic data volume and generation conditions on task performance with \sys. In our semi-supervised learning framework, the backbone model functions as a feature extractor, providing outputs that are used by task classifier \( f \) and per-class inlier-outlier detector \( g^j \).

\begin{figure}[t]
	\centering
    \begin{subfigure}{.95\linewidth}
	\includegraphics[width=\linewidth]{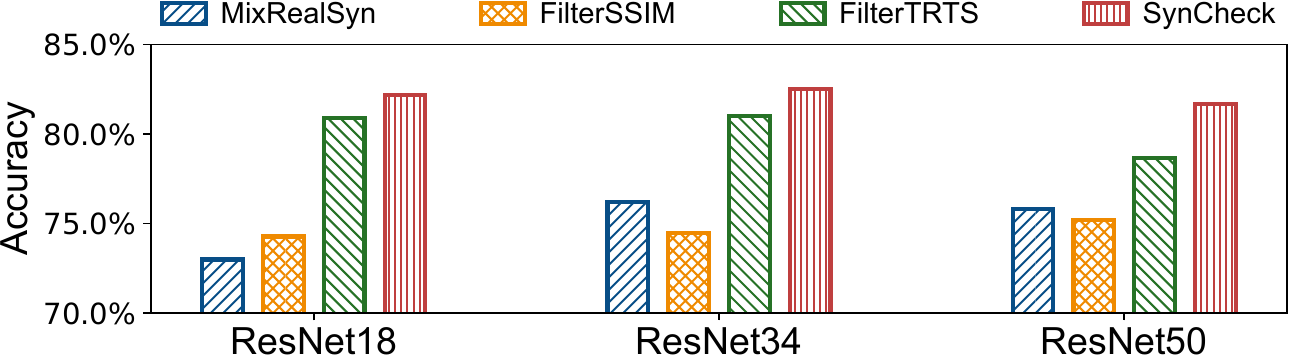}
    \vspace{-5mm}
	\caption{Cross-domain}
	\label{fig:cross_domain_backbone}
	\end{subfigure}
    \begin{subfigure}{.95\linewidth}
	\includegraphics[width=\linewidth]{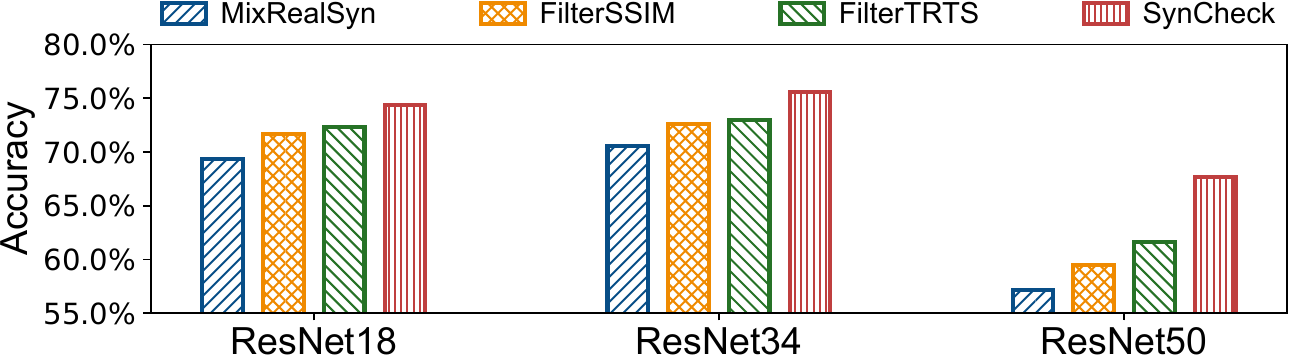}
	\vspace{-5mm}
    \caption{In-domain}
	\label{fig:in_domain_backbone}
	\end{subfigure}
    \vspace{-4mm}
	\caption{Performance v.s. Backbone Models.}
    \vspace{-5mm}
	\label{fig:backbone}
\end{figure}

\nosection{Impact of Backbone Model}
\label{sec:backbone_impact}
The results for different backbone models are presented in \figref{fig:backbone}. \sys consistently achieves superior task performance across all architectures, demonstrating its robustness and adaptability. Furthermore, the performance variations with RF-Diffusion highlight that increasing model parameters does not guarantee improved performance and may instead increase the risk of overfitting.


\begin{figure}[t]
	\centering
    \begin{subfigure}{.95\linewidth}
	\includegraphics[width=\linewidth]{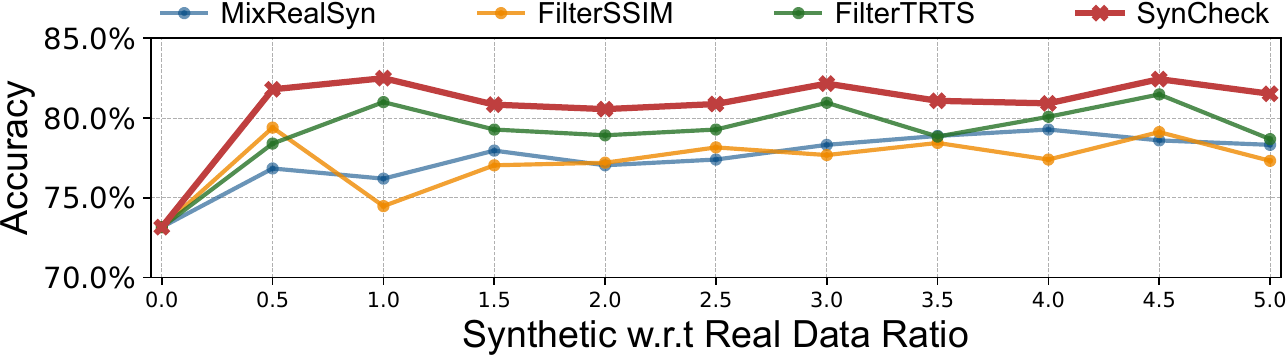}
    \vspace{-5mm}
    \caption{Cross-domain}
		\label{fig:cross_domain_amount}
	\end{subfigure}
    \begin{subfigure}{.95\linewidth}
	\includegraphics[width=\linewidth]{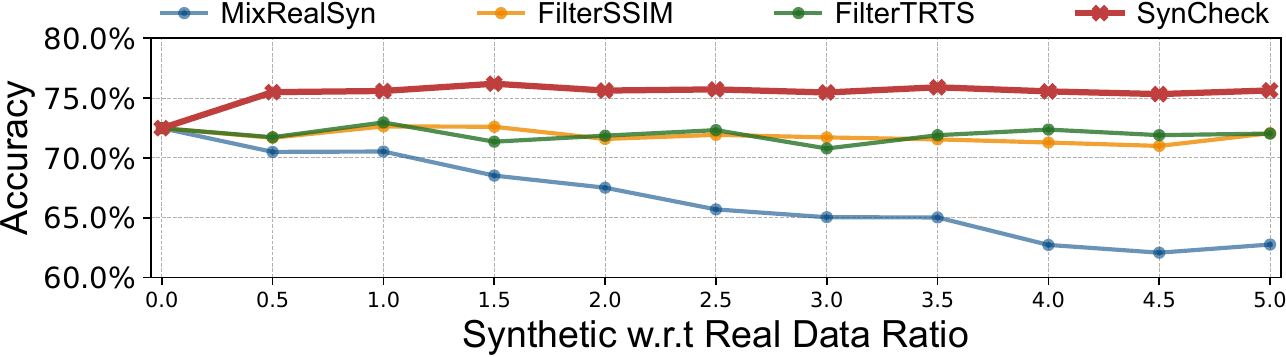}
    \vspace{-5mm}
    \caption{In-domain}
		\label{fig:in_domain_amount}
	\end{subfigure}
    \vspace{-2mm}
	\caption{Performance v.s. Synthetic Data Amount.}
    \vspace{-4mm}
	\label{fig:amount}
\end{figure}

\nosection{Impact of Synthetic Data Amount}
\label{sec:amount_impact}
We investigate the impact of synthetic data volume on task performance, as illustrated in \figref{fig:amount},by varying the initial synthetic data amounts from 0\% to 500\% of the real data in increments of 50\%. We compare \sys with the three synthetic data utilization methods employed in the overall performance evaluation. \sys consistently outperforms all three baseline methods across varying volumes of synthetic data. For instance, with 50\% synthetic data, \sys achieves performance metrics of 81.8\% and 75.5\% for in-domain and cross-domain setups, respectively, compared to 76.8\% and 70.5\% achieved by the nonselective mixture approach. This corresponds to a significant performance improvement of 6.5\% and 7.1\%, respectively. In the in-domain setup with 500\% synthetic data, the nonselective mixture approach results in a 13.4\% performance degradation compared to using real data alone, whereas \sys demonstrates a 4.3\% performance improvement. Overall, filtering synthetic data with TRTS using the task model yields superior results compared to filtering based on SSIM, highlighting the advantage of task-oriented filtering over methods relying on visual similarity metrics.

As the volume of synthetic data increases, the performance of the nonselective mixture approach exhibits distinct trends in cross-domain and in-domain setups. In the cross-domain scenario, the performance of the nonselective mixture generally improves, primarily due to the increased diversity of the trained classes. However, this performance gain saturates once the synthetic data exceeds 150\% of the real data volume, indicating that the diversity benefits of synthetic data reach a point of diminishing returns. On the contrary, in the in-domain setup, the nonselective mixture approach results in consistent performance degradation. This suggests that the affinity limitation cannot be mitigated by increased diversity, leading to train set distribution shifts from over-reliance on synthetic data.

In contrast, methods incorporating synthetic data filtering do not experience performance degradation with increased synthetic data volume, as they ensure the selection of data with better affinity. The gradual increase in synthetic data enhances diversity, contributing to further improvements in task performance. However, the rate of performance improvement begins to slow once the synthetic data volume exceeds 100\% of the real data.

\begin{figure}[t]
    \centering
    \includegraphics[width=.85\linewidth]{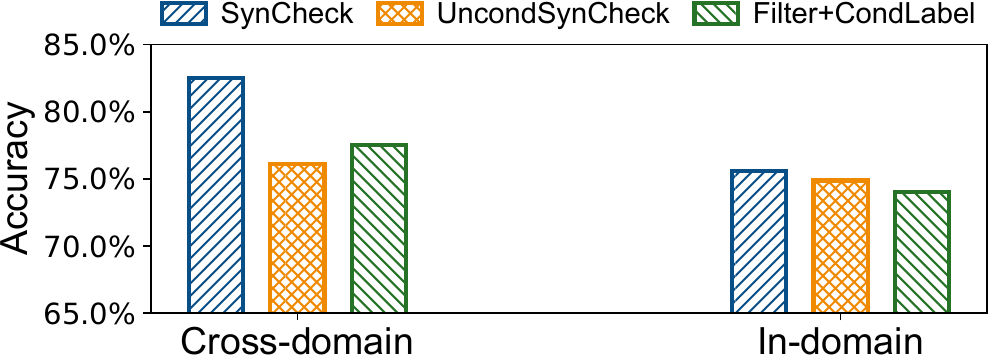}
    \vspace{-2mm}
    \caption{Condition assessment: pseudo-label origin and generation control.}
    \vspace{-3mm}
    \label{fig:cond_eval}
\end{figure}

\nosection{Impact of Pseudo-label Origin}
We assess the task performance of filtering synthetic data while utilizing generation conditions as labels. Specifically, we maintain the data filtering mechanism of SynCheck but replace the assigned pseudo-labels with generation conditions as labels during the training of the task model, referred to as `Filter+CondLabel' in \figref{fig:cond_eval}. Our evaluation reveals that replacing inlier labels with generation conditions leads to a decline in performance compared to the original \sys. This performance degradation can be attributed to a dual loss of affinity and diversity: the loss of affinity arises from the use of generation conditions as labels, while the loss of diversity results from the reduced volume of synthetic data after filtering. The observed decline underscores the critical importance of preserving both affinity and diversity in synthetic data utilization for optimal task performance.

\begin{tcolorbox}[size=small]
\textbf{Answer to RQ5}: Each design component in \sys contributes to task performance enhancement in the semi-supervised learning framework. While other factors, including backbone model and volume of initial synthetic data, affect the absolute performance, \sys consistently outperforms the nonselective mixture and other data selection methods. 
\end{tcolorbox}

\nosection{Necessity of Generation Conditions}
To underscore the significance of generation conditions, we replace conditional generative models with unconditional ones. The task performance using synthetic data generated from unconditional models, denoted as `Uncond\sys' in \figref{fig:cond_eval}, exhibits a decline compared to its conditional counterpart. This performance degradation emphasizes the necessity of generation conditions in synthetic data quality. We attribute this decline to the absence of generation control~\cite{graikos2023conditional}. The condition-based control ensures that generative models produce structured and meaningful samples. In contrast, unconditional models, lacking such constraints, inefficiently explore the data space~\cite{pandey2021vaes} during both training and inference, leading to the generation of lower-quality synthetic data.

The aforementioned experiments underscore the significance of conditions in both the data synthesis and data utilization processes. However, the cross-domain setup exhibits more pronounced performance degradation compared to the in-domain setup. We attribute this variation in performance to the data synthesis methodology employed by the selected in-domain generative model, RF-Diffusion. Specifically, RF-Diffusion operates by purifying noisy real-world data, where the data itself, apart from generation conditions, provides more informative cues than the cross-domain synthesis method utilizing a GAN model.

\begin{tcolorbox}[size=small]
\textbf{Answer to RQ6}: While wireless synthetic data from conditional generative models exhibits limited affinity, incorporating conditions during generative models' training is essential to ensure controllable and beneficial diversity.
\end{tcolorbox}

\begin{table}[t]
    \caption{Per-batch Training Overhead}
    \vspace{-3mm}
    \centering
    \label{tab:overhead_cmp}
    \begin{adjustbox}{max width=\linewidth}
    \begin{tabular}{cccc}
    \toprule
    \textbf{Methods} & \textbf{Parameters (M)} &\textbf{FLOPs (G)} & \textbf{GPU memory (MB)} \\
    \midrule
    Standard Model & 7.58 & 15.2491 & 337.96  \\
    \sys  & 7.59 & 15.2495 & 606.42  \\
    \bottomrule
    \end{tabular}
    \end{adjustbox}
    \vspace{-3mm}
\end{table}

\subsection{Computational Overhead Analysis} 
We evaluate the computational overhead of \sys in comparison to baseline methods. The additional costs incurred by \sys arise from two main factors: (1) the increased parameter amount due to per-class inlier-outlier detectors, and (2) the additional forward passes required for synthetic data with domain-specific processing to enable pseudo-label supervision. These factors only impact the training phase, as inference in \sys relies solely on the task classifier and does not involve inlier-outlier detection or pseudo-label assignment.
In contrast, all baseline methods employ standard task training procedures. When filtering is applied (\eg, SSIM or TRTS), synthetic data is selected before training. As a result, the per-batch computational cost remains consistent across all baselines. To quantify the overhead, we compare \sys and the baselines in three aspects: parameter count, FLOPs (floating-point operations), and GPU memory consumption. The results are presented in \tabref{tab:overhead_cmp}. It turns out that \sys introduces a 0.13\% growth in parameters and a negligible 0.0026\% rise in FLOPs. However, GPU memory usage during training nearly doubles compared to standard training. This overhead is incurred only during training and can be amortized during inference, making the additional resource requirement acceptable given the accuracy improvement.

\section{Discussion}
\nosection{Data Quality Metrics} 
We adopt the principles of using task performance as the ultimate standard for data quality~\cite{affinitydiversity} and assessing quality on two key attributes: affinity and diversity. Our quality metrics have demonstrated their effectiveness through correlations with target task performance. While prior work identifies another attribute of \textit{generalization}~\cite{faithful}, it stresses no direct copy of real data for privacy issues~\cite{jordon2021hide} and beyond the scope of our paper.


 \nosection{Trade-offs between Affinity and Diversity} Affinity and diversity are interdependent attributes of synthetic data, where improving one may compromise the other. In this paper, we focus on enhancing affinity while preserving diversity to improve task performance compared to the quality-oblivious utilization. The optimal balance between the two quality metrics is left for future work.

\nosection{Empirical Analysis Scale}
Our approach to selecting representative generative models balances two criteria: (1) generation techniques, which encompass the widely adopted frameworks of GAN (implicit likelihood estimation~\cite{grover2018flow}), and diffusion models (explicit likelihood estimation~\cite{pandey2023complete}); and (2) generation domains, which include both cross-domain and in-domain configurations to ensure coverage of diverse data generation scenarios. Additionally, to maintain evaluation fidelity and reproducibility, we prioritize models with open-source implementations, enabling fair and consistent comparison of our data utilization methods. 


\nosection{Better Conditional Generative Models} Given the deficiencies of existing wireless generative models, an alternative solution is to develop improved generative models tailored to the wireless domain. This approach is orthogonal to our utilization scheme and can be seamlessly integrated to further enhance task performance. 

\nosection{Availability of High-quality Synthetic Data}
The efficacy of our data selection framework relies on generative models' ability to produce synthetic data with high affinity. Our experiments show that modern generative models can reliably produce high-affinity samples with enough sampling attempts.

\nosection{Applicability to Other Tasks and Modalities}  
While \sys primarily focuses on classification tasks within the WiFi domain, there have been efforts to integrate synthetic data with real data for non-classification tasks and other modalities, such as channel prediction~\cite{newrf} and mmWave applications~\cite{rf-genesis}. For regression tasks, techniques from anomaly detection in time-series data~\cite{blazquez2021review} could be adapted for inlier-outlier detection. The detection aligns with existing anomaly detection methods that map data to a unified space and compute anomaly scores~\cite{dlanomaly}. For other modalities, since our framework’s design is not inherently tied to specific signal properties, \sys can be extended by adapting the domain-specific processing in the label assignment phase. We leave the exploration of these directions to future work.


\section{Related Work}

\nosection{Data-driven Generative Models}
We focus on evaluating the quality of wireless synthetic data generated by data-driven generative models and propose a taxonomy that categorizes these models into three groups based on the source and target data domains, as detailed in \secref{sec:literature}. In addition to the representative models considered in our assessment, other relevant work in cross-domain transfer includes DFAR~\cite{wang2020cross}, which aligns cross-domain samples using decision boundaries and centers, and SCTRFL~\cite{SCTRFL} as well as AutoFi~\cite{AutoFi}, which calibrates samples distorted by temporal environmental changes using autoencoders. In-domain generative models include RF-EATS~\cite{rf-eats} amplifying liquid-related measurements using variational autoencoders, and AF-DCGAN~\cite{AF-DCGAN} extending localization fingerprints.
Although these works use closed-source models and datasets, which complicate the fair evaluation of original systems, their design rationales align with representative models in our study. Therefore, we argue that they can also benefit from quality-guided utilization for better performance.

\nosection{Physics-Informed Generative Models} In addition to data-driven generative models, another category includes physics-informed models, which integrate physical laws during data generation. The differences in their mechanisms mean our current framework cannot be directly applied to purely physics-based models. However, existing works on physical models tend to combine with data-driven models~\cite{rf-genesis,rf-eats} for synthetic data refinement. This partial reliance on data-driven elements suggests potential applicability of our method,  and we leave this exploration for future work.

\nosection{Verified Synthetic Data for Enhanced Performance}
While synthetic data has been demonstrated to risk performance degradation due to distributional shifts~\cite{NeurIPS2024_MC,Dohmatob2024ATO}, recent advances highlight verification as a critical mitigation strategy~\cite{Feng2024BeyondMC}. By selectively filtering synthetic data, even imperfect generative models can yield better performance. Our work aligns to the important quality-aware data selection mechanism, and extends by using downstream task performance as a measure of real-synthetic distribution alignment. We further introduce iterative filtering during training to dynamically refine synthetic data quality.

\nosection{Semi-supervised Learning}
Semi-supervised learning leverages both labeled and unlabeled data for model training. Key techniques include decision consistency regularization across different augmented versions of the same sample~\cite{laine2016temporal} and pseudo-label assignment for unlabeled data~\cite{pseudo_label}. While our semi-supervised framework builds on these established methods, it uniquely treats synthetic data as unlabeled and selects a useful subset based on its quality. This approach offers a plug-and-play solution to address synthetic data quality issues, distinguishing it from most prior semi-supervised learning that focuses solely on real data.

\section{Conclusion}
While synthetic data complements the data quantity for the wireless community, its quality has often been overlooked, limiting its ability to enhance task performance. In this paper, we introduce tractable and generalizable quantification metrics for two key data quality attributes—affinity and diversity—to systematically evaluate the quality of existing wireless synthetic data. Based on insights derived from our quality assessment, we propose a universal utilization scheme that enhances data affinity while preserving diversity, leading to improved task performance. We anticipate that the quantification of affinity and diversity will become a standard practice in future evaluations of wireless synthetic data. Additionally, our quality-aware utilization framework can be seamlessly integrated with various wireless generative models to boost their performance.


\appendix
\section{Theorem Proof}\label{theorem_proof}

\begin{lemma} Given the task of predicting label $y$ from sample $x$, the performance optimization is achieved through training a classifier $\hat{y}=f(x)$ to minimize the loss~\cite{cas} 
\begin{equation}
    \mathbb{E}_{p(x,y)}[\mathcal{L}(y,\hat{y})]=\mathbb{E}_{p(x)}[\mathbb{E}_{p(y|x)}[\mathcal{L}(y,\hat{y}\mid X=x)]]
\end{equation}
During training with concrete samples, we empirically optimize the loss $\frac{1}{N}\mathcal{L}(y,\hat{y})$ and the optimal prediction is: 
\begin{equation}\label{eqn:loss}
\hat{y}=\arg\min \mathbb{E}_{p(y|x)}[\mathcal{L}(y,\hat{y}\mid X=x)]
\end{equation}
\end{lemma}

\begin{definition}[Total Variation] Consider $p,q$ probabilities on $E$, the total variation ($\TV$) distance~\cite{tv} between $p$ and $q$ is $\TV(p,q)=\sup_{A\subset E}|p(A)-q(A)|=\frac{1}{2}\int \left|p(x)-q(x)\right| dx$. 
\end{definition}

\begin{proof}[\textbf{Proof of Theorem 1}]
Based on \eqnref{eqn:loss}, given the training data distribution \( p(x, y) \), the test loss under the data distribution \( p_{\theta}(x, y) \) is expressed as:
\[
\mathbb{E}_{TRTS}[L(y, \hat{y})] = \mathbb{E}_{p_{\theta}(x)}\left[\mathbb{E}_{p(y|x)}[L(y, \hat{y} \mid X = x)]\right].
\]
The marginal distributions of synthetic and real samples are:
\[
p_{\theta}(x) = \int_y p_{\theta}(x, y) \, dy, \quad p(x) = \int_y p(x, y) \, dy.
\]
For any measurable set \( A \), the probability difference satisfies:
\[
|p_\theta(x \in A) - p(x \in A)| \leq \int_y |p_\theta(x|y) - p(x|y)| \, p(y) \, dy,
\]
which implies proximity in the total variation (TV) distance:
\[
\text{TV}(p_\theta, p) \leq \mathbb{E}_{p(y)}\left[\text{TV}(p_\theta(x|y), p(x|y))\right].
\]
Thus, similarity in the conditional distributions ensures similarity in the marginal distributions. Let \( h(x) = \mathbb{E}_{p(y|x)}[L(y, \hat{y} \mid X = x)] \). The difference between the training and testing losses is given by:
\[
|\mathcal{L}_{\text{test}} - \mathcal{L}_{\text{train}}| = \left|\mathbb{E}_{p_\theta(x)}[h(x)] - \mathbb{E}_{p(x)}[h(x)]\right|.
\]
If \( h(x) \) is bounded (e.g., \( \mathcal{L} \in [0, 1] \)), it follows that:
\[
\begin{aligned}
|\mathbb{E}_{p_\theta}[h] - \mathbb{E}_p[h]| & \leq 2 \cdot \text{TV}(p_\theta, p) \cdot \sup |h(x)| \\
& \leq 2 \cdot \mathbb{E}_{p(y)}\left[\text{TV}(p_\theta(x|y), p(x|y))\right] \cdot \sup |h(x)|.
\end{aligned}
\]
This implies that the test loss \( \mathcal{L}_{\text{test}} \) approximates the training loss \( \mathcal{L}_{\text{train}} \). In summary, under the sufficient condition that the synthetic data distribution \( p_\theta(x|y) \) is close to the real data distribution \( p(x|y) \), a model achieving a low training loss will also achieve a low testing loss on synthetic data.

\end{proof}

\begin{proof}[\textbf{Proof of Theorem 2}]
With training distribution $p_{\theta}(x,y)$, the test loss under the data distribution $p(x,y)$ is given by: 
\[
    \mathbb{E}_{TSTR}[L(y,\hat{y})]=\mathbb{E}_{p(x)}[\mathbb{E}_{p_{\theta}(y|x)}[\mathcal{L}(y,\hat{y}\mid X=x)]]
\]

Under the sufficient condition that $p_{\theta}(y|x)\approx p(y|x)$, the test loss on real data becomes: 
\[
\mathbb{E}_{p(x)}[\mathbb{E}_{p(y|x)}[\mathcal{L}(y,\hat{y}\mid X=x)]].
\]
This aligns with training loss on real data, resulting in low test loss. Therefore, consistency between $p_{\theta}(y|x)$ and $p(y|x)$ ensures good test performance in train-synthetic setup.
\end{proof}

\section*{Acknowledgments}
We sincerely thank the anonymous shepherd and reviewers for their insightful critique and constructive feedback, which have significantly improved the quality of this paper. This work is supported by National Key R\&D Program of China (Grant No.2023YFF0725004) and National Natural Science Foundation of China (Grant No.62272010 and 62061146001). Chenren Xu (chenren@pku.edu.cn) is the corresponding author.

\ifeg
TBD
\else
\fi

\bibliographystyle{unsrt} 
\bibliography{ref}

\end{document}